\documentclass[times,twocolumn,final]{elsarticle}
\usepackage{medima}
\usepackage{framed,multirow}

\usepackage{amssymb}
\usepackage{latexsym}

\usepackage{url}
\usepackage{xcolor}

\definecolor{newcolor}{rgb}{.8,.349,.1}

\journal{Medical Image Analysis}

\usepackage{amsmath, amssymb,bm}
\usepackage{booktabs}
\usepackage{graphicx,epstopdf}
\usepackage{multirow}
\usepackage{tabularx,threeparttable}
\usepackage{array}
\usepackage{comment}
\usepackage[hidelinks]{hyperref}
\usepackage{lineno}
\usepackage{makecell}
\usepackage{ulem}
\usepackage{mathrsfs}
\usepackage{caption}
\usepackage{amssymb}
\usepackage{graphicx} 
\usepackage{caption}
\usepackage{subcaption}

\newcommand{\para}[1]{\vspace{.05in}\noindent\textbf{#1}}
\def\eg{e.g.}
\def\ie{\textit{i.e.}}

\newcommand\T{\rule{0pt}{2.6ex}}       
\newcommand\B{\rule[-1.2ex]{0pt}{0pt}} 
\newcolumntype{?}[1]{!{\vrule width #1}}

\usepackage{numcompress}
\biboptions{authoryear}

\begin{document}

\begin{frontmatter}

\title{RMDL: Recalibrated Multi-instance Deep Learning  for Whole Slide Gastric Image Classification}
\address[CUHK]{Department of Computer Science and Engineering, The Chinese University of Hong Kong, Hong Kong, China}
\address[ZS6Y]{{Department of Pathology, the Sixth Affiliated Hospital of Sun Yat-sen University, China}}
\address[ImsightMed]{Imsight Medical Technology Co., Ltd., China}
\address[ZS6Y2]{{Department of Radiation Oncology, the Sixth Affiliated Hospital of Sun Yat-sen University, China}}

\cortext[Co-firstauthor]{{Co-first authors.}}
\cortext[mycorrespondingauthor]{{Corresponding authors. (fanxjuan@mail.sysu.edu.cn, hchen@cse.cuhk.edu.hk)}}

\author[CUHK]{Shujun Wang\corref{Co-firstauthor}}
\author[ZS6Y]{{Yaxi Zhu}\corref{Co-firstauthor}}
\author[CUHK]{Lequan Yu}
\author[ImsightMed]{Hao Chen\corref{mycorrespondingauthor}}
\author[CUHK,ImsightMed]{Huangjing Lin} 
\author[ZS6Y2]{{Xiangbo Wan}}
\author[ZS6Y]{{Xinjuan Fan}\corref{mycorrespondingauthor}}
\author[CUHK]{Pheng-Ann Heng}

\begin{abstract}
	The whole slide histopathology images (WSIs) play a critical role in gastric cancer diagnosis.
	However, due to the large scale of WSIs and various sizes of the abnormal area, how to select informative regions and analyze them are quite challenging during the automatic diagnosis process. 
	The multi-instance learning based on the most discriminative instances can be of great benefit for whole slide gastric image diagnosis.
	In this paper, we design a recalibrated multi-instance deep learning method (RMDL) to address this challenging problem.
	We first select the discriminative instances, and then utilize these instances to diagnose diseases based on the proposed RMDL approach.
	The designed RMDL network is capable of capturing instance-wise dependencies and recalibrating instance features according to the importance coefficient learned from the fused features.
	Furthermore, we build a large whole-slide gastric histopathology image dataset with detailed pixel-level annotations.
	Experimental results on the constructed gastric dataset demonstrate the significant improvement on the accuracy of our proposed framework compared with other state-of-the-art multi-instance learning methods.
	Moreover, our method is general and can be extended to other diagnosis tasks of different cancer types based on WSIs.	

\end{abstract}

\end{frontmatter}


\section{Introduction}
\label{sec:introduction}
\begin{figure*}[t]
	\centering
	\includegraphics[width=0.65\linewidth]{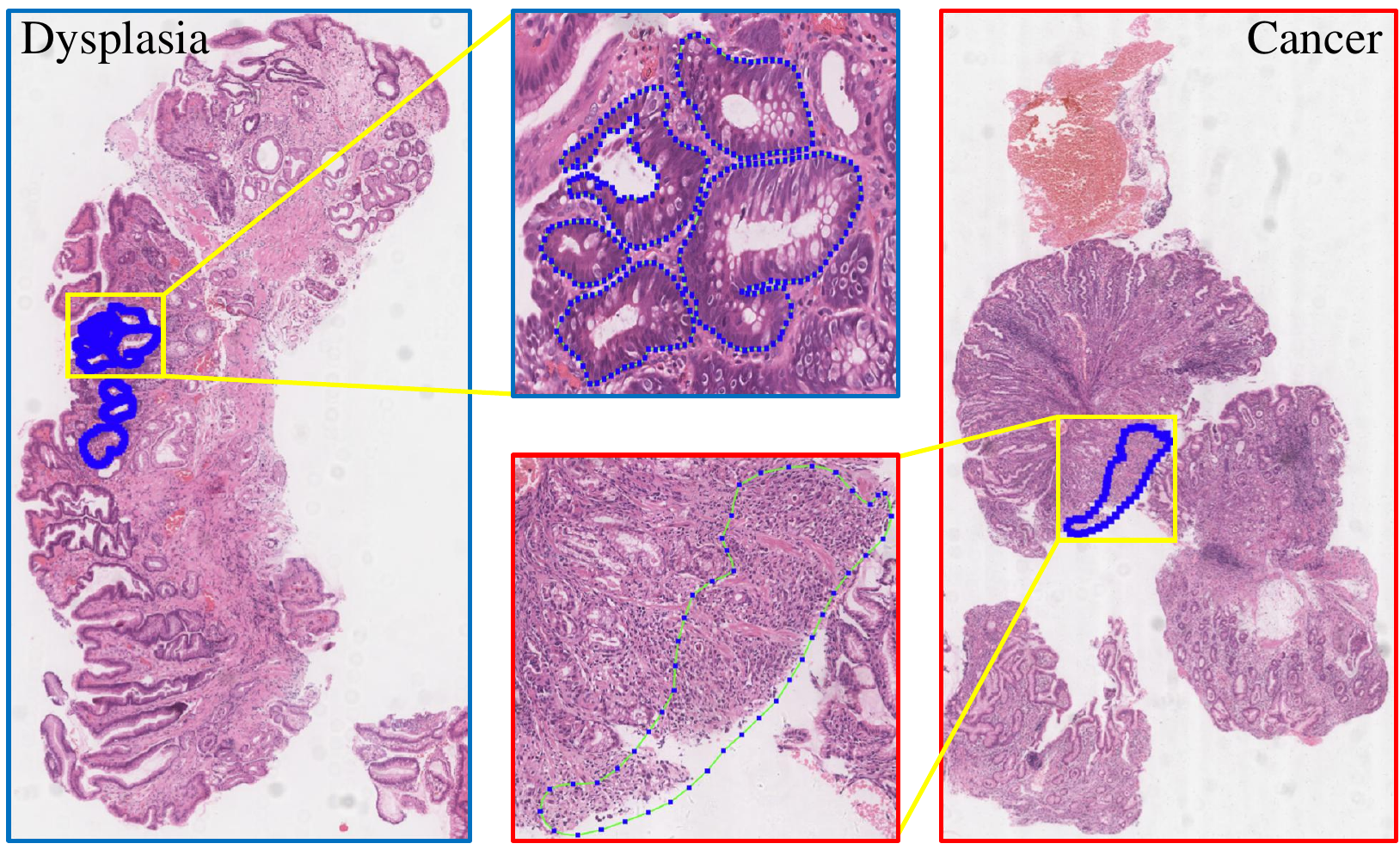}
	\caption{Illustration of whole slide gastric histopathology images. The left and right columns represents the images with label of \textit{Dysplasia} and \textit{Cancer}, respectively. The abnormal regions are annotated with blue dots and zoomed in the middle column.}
	\label{fig:background}
\end{figure*}

{
	According to global statistics from the world health organization\footnote{\url{http://gco.iarc.fr/today/home/}}, gastric cancer is the third most common cause of cancer-related mortality.}
Gastric cancer has been considered as a single heterogenous disease with several histopathologic characteristics~\citep{shah2011molecular}.
And histopathological diagnosis is extremely critical for definitive and supportive reference results~\citep{yasui2001molecular} in clinical.
With the emergence of whole slide imaging technology, the whole slide histopathology image (WSI) analysis has been accelerated~\citep{sertel2009computer}.
The manual pathological analysis by traversing the entire WSI with diverse magnifications  is subjective and time-consuming resulting from the large scale of WSIs (typically 100,000 $\times$ 100,000 pixels).
Therefore, the automated and accurate analysis of WSIs is promising in improving diagnostics, predicting growth trends, and designing treatment strategies~\citep{fonseca2017lauren,van2017prognostic}.
In gastric cancer classification, except for the normal and cancer types, dysplasia defined as a benign neoplastic lesion is highly associated with cancer area in whole slide gastric histopathology images~\citep{li2016risks}.
{In this paper, we aim to develop an automatic gastric cancer classification method for the WSI into normal, dysplasia, and cancer types.
}
The challenges of the automatic gastric cancer classification are threefold:
1) the large intra-class and small inter-class variations on texture and morphology of histopathology patches make diagnosis process ambiguous, as shown in Figure~\ref{fig:background}; 
2) due to the considerable scale and high computing requirements of WSI, it is difficult to process the entire WSI at once; 
and 3) the discriminative information of WSI is prone to be suppressed when the abnormal regions only occupy a small proportion of the whole slide image compared with the normal regions in whole slide image (\eg~Figure~\ref{fig:background}).

{With the benefits of convolutional neural network (CNN), automated histopathology image analysis has achieved promising performance in mitosis detection \citep{cirecsan2013mitosis,veta2015assessment, albarqouni2016aggnet}, gland instance segmentation~\citep{sirinukunwattana2017gland,graham2018mild}, cell detection~\citep{wang2016subtype}, nuclei analysis~\citep{xie2015beyond,xing2016automatic,sirinukunwattana2016locality,xu2016stacked,zhou2017sfcn}, metastasis  detection~\citep{kong2017cancer,liu2017detecting,bejnordi2017diagnostic,lin2018scannet,lin2019fast} and histopathology image classification~\citep{bayramoglu2016deep, bug2016multi,bandi2019detection}.}
For example, \cite{chen2016mitosis} presented a fast and accurate method to detect mitosis via a deep cascaded neural network. 
And \cite{bentaieb2016topology} trained a deep network to encode geometric and topological priors for histology gland segmentation.
{However, most of those CNN-based methods focused on the pre-extracted regions of interests (ROIs) from WSIs, since the enormous size of WSI and the limitation of computing resources may hinder the direct process of WSI.}

Currently, some approaches were proposed to conduct whole slide image analysis~\citep{xu2017large,zhu2017wsisa,bentaieb2018predicting,ren2018adversarial}.
{
	For example, \cite{zhu2017wsisa} proposed a method for survival prediction from WSIs, which consists of adaptive patch sampling, patch clustering, cluster-level Deep Convolutional Survival (DeepConvSurv) prediction, and patient-level aggregation.
	A recurrent visual attention model for WSI cancer localization was presented by~\cite{bentaieb2018predicting}.}
A domain adaptation method based on Siamese architecture for the prostate WSI classification between two different datasets {was designed} by~\cite{ren2018adversarial}.
{However, it is challenging for these methods {to be applied} on the disproportion dataset which has a small proportion of abnormal regions, especially for gastric cancer classification {task, since features} of normal regions would suppress the discriminative information {from abnormal tissues.}
}

To conduct the WSI classification for the disproportion dataset, two non-trivial points need to be studied carefully: 1) selecting the informative patches (ROIs), and 2) aggregating the patch-level information effectively.
{Generally, among the abnormal and normal patches extracted from one whole slide image,} the reasonable image label prediction should be made by considering the contrast of abnormal and normal information and focusing on the most discriminative features.
{For instance, in the gastric WSI classification task, cancer patches are more critical than normal ones, when classifying a WSI into cancer type.}
\cite{Hou} presented an expectation maximization (EM) based procedure to locate discriminative patches before image-level prediction.
However, it is infeasible to apply this EM-based method on the disproportion dataset in clinical practice, due to the small proportion of the abnormal area and the huge computation cost of training and inference on each iteration.
{
	What's more,
	\cite{mercan2018multi} extracted candidate ROIs according to specific pathologist viewing logs.}
However, this procedure is objective and difficult to generalize on the gastric WSI classification task. 
Therefore, it still needs to be investigated about how to locate the discriminative regions for the gastric WSI classification effectively.
{Some previous methods formulated the procedure of aggregating patch-level features into image-level predictions as multi-instance learning (MIL) problem, where each patch represents one instance, and each WSI is regarded as a bag~\citep{dundar2010multiple,quellec2017multiple,zhu2017deep}.}
Based on this formulation, various methods were proposed to solve the whole slide histopathology image analysis problems~\citep{xu2014deep,courtiol2018classification,ilse2018attention}.
For example, \cite{xu2014deep} utilized CNNs to extract features of high-resolution images and MIL-Boost to classify high-resolution histopathology images.
\cite{li2015multiple} presented a MIL based boosting regularized tree to tackle with cancer detection from histopathology images.
A Noise-AND pooling method was developed to classify and segment microscopy images in~\cite{kraus2016classifying}.
\cite{zhu2017deep} proposed an end-to-end trained deep sparse multi-instance pooling method for whole mammogram classification.
Recently, attention-based multi-instance learning methods appeared in medical image analysis community.
\cite{courtiol2018classification} designed a method to consider top instances and negative ones simultaneously to cope with histopathology classification and localization tasks.
{\cite{ilse2018attention} presented an attention-based MIL pooling mechanism by combining with neural networks to solve natural image and histopathology image analysis problems. }
{
	However, these methods treat all instances independently and ignore the interrelated association among different instances.}
Without considering the interrelated information among different instances, the attention-based methods are difficult to estimate the contribution of instances and make accurate image label predictions.
\begin{figure*}[t]
	\centering
	\includegraphics[width=0.8\linewidth]{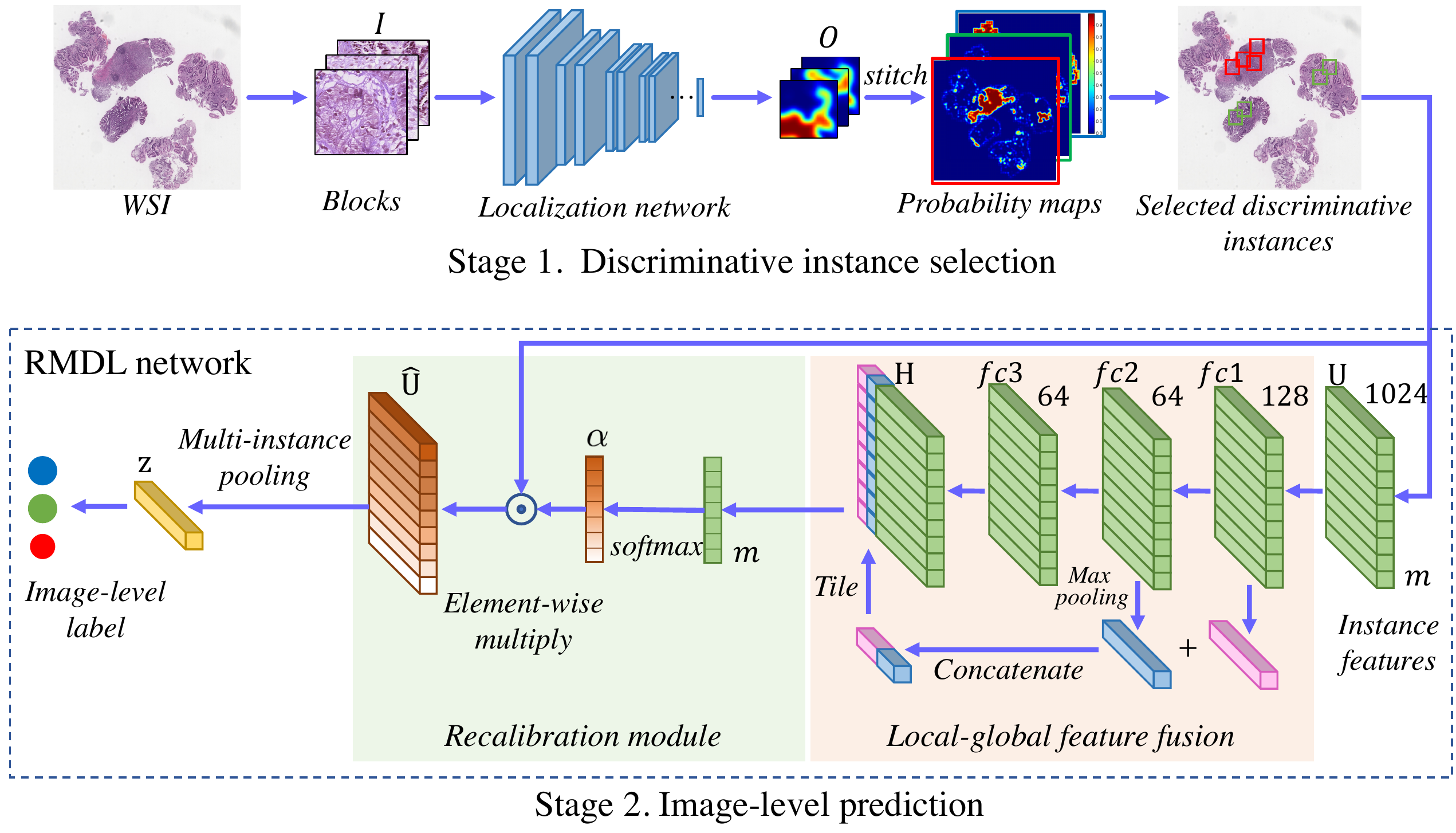}
	\caption{
		The flowchart of our proposed method (Best viewed in color). In the first stage, a localization network is trained to select the discriminative instances.
		In the second stage, we design an RMDL network for image-level label prediction, which consists of local-global feature fusion, instance recalibration, and multi-instance pooling modules.}
	\label{fig:whole_architecture}
\end{figure*}

In this paper, we present an efficient two-stage framework for the automated gastric WSI classification.
As shown in Figure~\ref{fig:whole_architecture}, our framework consists of two successive stages: discriminative instance selection and image-level prediction.
We first extract the discriminative patches based on the predicted probability maps generating from a localization network.
Particularly, we design a patch-based fully-convolutional localization network to speed up the detection process.
Then, we develop a novel recalibrated multi-instance deep learning network (RMDL) for image-level classification to overcome the drawback of directly aggregating the prediction of discriminative patches.
The novel recalibrated multi-instance deep learning module is able to capture the instance-wise dependencies and recalibrate instance features according to the importance coefficient learned from the combined features.
{Compared with the traditional multi-instance learning methods, the proposed RMDL provides an effective option to explore the interrelationship of different patches and consider the various impact of them to image-level label classification.}
The designed RMDL network sheds light on the large scale image analysis problems.
We evaluate the proposed framework on our own collected large gastric WSI dataset.
{We conduct extensive analysis of our method and demonstrate that our proposed RMDL outperforms other multi-instance learning methods.}
Overall, the main contributions of our work are summarized as follows.
\begin{enumerate}
	
	\item We present an efficient two-stage framework including discriminative instance selection and recalibrated multi-instance deep learning (RMDL), for the gastric whole slide image classification.
	
	\item {The developed novel RMDL network considers different contributions of patches to predict the final image-level label by recalibrating instance features automatically.}
	
	\item We construct a large whole slide gastric image dataset which consists of 608 images. The extensive experiments on the dataset demonstrate the effectiveness of our RMDL network and our method outperforms other multi-instance learning methods.
	

\end{enumerate}

The remainders of this paper are organized as follows.
We first introduce the WSI dataset we collected in Section~\ref{sec:dataset}. 
We elaborate the details of our framework in Section~\ref{sec:method}. 
The experiments and analysis are presented in Section~\ref{sec:experiments}.
We further discuss our method in Section~\ref{sec:discussion} and conclusions are drawn in Section~\ref{sec:conclusions}.

\section{Dataset}
\label{sec:dataset}
We constructed a new large Whole Slide Gastric Image dataset (WSGI) by collaborating with The Sixth Affiliated Hospital of Sun Yat-sen University.
The WSGI dataset consists of 608 whole slide images and {all the images are collected from different patients.}
These slides are stained by Hematoxylin and Eosin and scanned by a Leica Aperio CS2 scanner at a 40X magnification with 0.2517 $\mu$m/pixel resolution.
They are stored in a multi-resolution pyramid structure containing multiple down-sampled images on different levels (the largest level is four), and the mean resolution of these slides is about $170,000 \times 80,000$ pixels.
Manual pixel-level annotations for our WSGI dataset are done by a pathologist with seven years of experience.
Each whole slide image is manually annotated with polygon outlines which encircle the abnormal (dysplasia and cancer) regions using the open-source Automated Slide Analysis Platform (ASAP).
These polygon annotations are stored as an ordered list of vertex (X, Y) pixel coordinates on level 0 and are saved as XML format files.
The image-level label includes three classes: (0) Normal, (1) Dysplasia, and (2) Cancer, where the number represents the abnormal grading.

There are 408 and 200 images in the training and testing parts according to the following principle. 
We randomly split the Dysplasia and Cancer images following a ratio of 1:4 for testing and training parts. Actually, the abnormal images also contain the normal regions and these regions can also be utilized to train the localization network and the all patches from normal images may be selected to train the network. And in clinical, the number of normal cases is also more than the number of abnormal ones. Therefore, we reduce the perception of normal images on the training part and increase the number of normal cases on the testing part.
The details of these two sets are shown in Table~\ref{table:data_statistic}.
For the dataset preprocessing, OTSU algorithm~\citep{otsu1979threshold} is used to extract tissue regions according to the gray value of the image. 
This operation effectively removes the irrelevant background of the images and speeds up the diagnostic process.

\begin{table}
	\begin{center}
		\caption{Statistic information of the dataset WSGI.}
		\label{table:data_statistic}
		\resizebox{0.45\textwidth}{!}
		{
		\begin{tabular}{c|ccc|ccc}
			\hline    
			&\multicolumn{3}{c|}{Training part} & \multicolumn{3}{c}{Testing part} \\
			\hline
			Class & Normal & Dysplasia & Cancer & Normal & Dysplasia & Cancer \\
			\hline
			${No.}$ & 23 & 139 & 246  &94 & 33 &  73  \\
			\hline
		\end{tabular}
	}
	\end{center}
\end{table}

\section{Methodology}
\label{sec:method}

Given a whole slide histopathology image $X$, our goal is to predict the image label $Y$ by considering extracted features of the discriminative patches \{$x_{1}, x_{2},...,x_{m}$\}. 
To this end, we design our whole slide gastric image classification framework in a two-stage way, as shown in Figure~\ref{fig:whole_architecture}.
We first extract the discriminative instances based on the abnormal probability map $P$ predicted by the localization network.
Then, we design a novel RMDL network to aggregate the candidate instance features into the image-level prediction $Y$ by recalibrating the importance coefficient $\alpha$ of each instance.

\subsection{Discriminative Instance Selection}
\label{sec:instanceselection}

For the WSI with small abnormal regions (\eg~Figure~\ref{fig:background}), it is necessary to extract the discriminative instances for further processing to avoid the suppression of a large proportion of the normal regions against the discriminative information.
%
%
Under these circumstances, there is no need to detect the accurate abnormal region contours but the coarse locations.
Therefore, we train a fully convolutional classification network (referred as \textit{localization network}) to detect abnormal regions following previous works~\citep{chen2016mitosis, lin2018scannet} instead of a  time- and memory-consuming segmentation network.
In this step, a patch-based mapping $f$ (localization network) is optimized to capture the discriminative information and predict the abnormal grading for each patch so that we obtain a probability map $P=f(X)$ for arbitrary WSI $X$.
%

\begin{figure*}[t]
	\centering
	\includegraphics[width=0.7\linewidth]{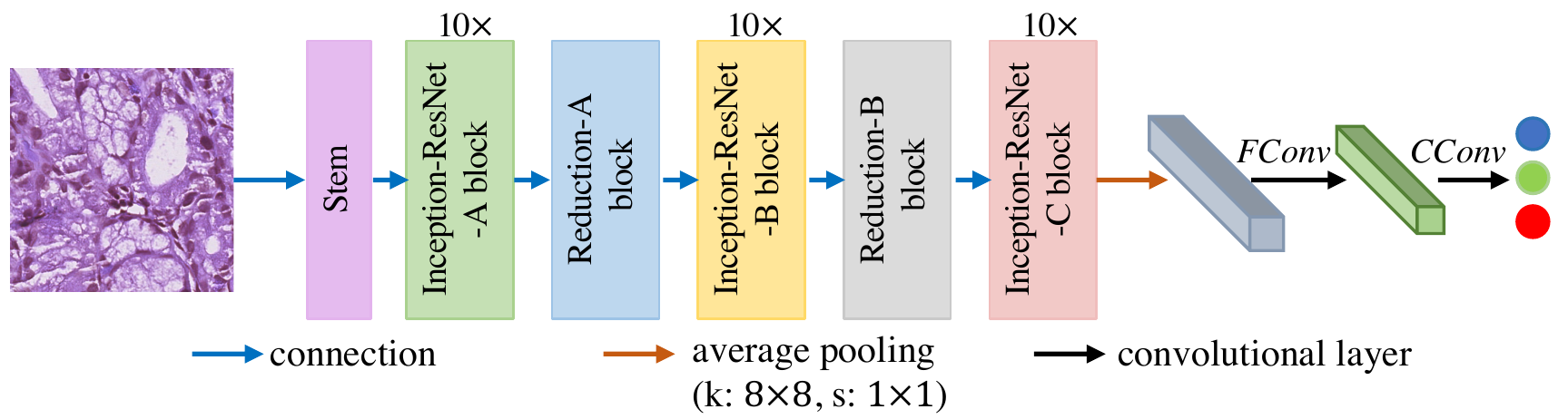}
	\caption{Localization network architecture. 
		We transfer the initial Inception Resnet\_v2 into a fully convolutional network by adding an Average Pooling layer, a convolutional layer for feature extraction (FConv), and another convolutional layer for classification (CConv). 
	}
	\centering
	\label{fig:resnet_architecture}
\end{figure*}

\para{Localization network.}
The localization network is the adaptive version of Inception Resnet\_v2 \citep{szegedy2017inception}.
To transfer the initial network into a fully convolutional one, we discard the final Global Average Pooling layer and the Fully connected layer for classification, then add an Average Pooling layer, a convolutional layer for feature extraction (FConv), and another convolutional layer for classification (CConv), as shown in Figure~\ref{fig:resnet_architecture}.
The added average pooling layer has a kernel of $8\times8$ and a stride of $1\times1$; the FConv and CConv have 1024 and 3 output channels, respectively.
Speeding up the network inference process is necessary to the histopathology image classification due to the large scale of WSIs.
Instead of predicting the probability of each patch, we divide the WSI into the large overlapped blocks and feed these blocks into the fully convolutional localization network to obtain a small probability map at once.
%
In particular, supposing that the sizes of one block and one patch are $I \times I$ and $N \times N$, respectively, the size of the output probability map ($O \times O$) can be calculated by $O={\lceil (I-N)}/{S} \rceil+1$, 
where $S$ is calculated by cumulative multiplication of the strides of down-sampling layers in the localization network ($S=32$ in our experiments).
The size of $I$ can balance the prediction speed and computation memory. 
We set $I$ as $1899$ so that the size of the generated probability map of each block is $51 \times 51$. 
And the size of block stride is carefully set as $1632$ according to $I-N+S$.
Therefore, we then obtain the probability maps of normal, dysplasia, and cancer types for each WSI by stitching the small probability map of each block without any overlap.

%
\para{Training patch selection.}
The pixel-level annotations are necessary to train a representative localization network for the WSI with a small proportion of abnormal regions.
In the context of disproportion dataset, assigning the abnormal regions only with image-level labels is quite imprecise due to the limited image-level supervision and diverse size of abnormal regions in histopathology images.
%
%
Therefore, we design a training patch selection method relying on abnormal region annotations to train the localization network.
For the abnormal WSIs, we extract the abnormal patches (cancer and dysplasia area) in an overlapped sliding window manner where the stride is set as $224$ and the normal patches are extracted randomly from the remaining regions.
For the rest of the normal WSIs, we randomly extract the normal patches from the whole slide regions.
After the training patch selection, the localization network can be optimized in a supervised way to capture the representative features through stacked convolutional layers. 

\para{Discriminative instance selection.}
We select the discriminative instances \{$x_{1}, x_{2},...,x_{m}$\} by the predicted probability map $P$ consisting of three channels for each WSI.
In detail, we adopt the non-maximum suppression (NMS) principle \citep{felzenszwalb2010object} to locate the top $m'$ discriminative patches based on the probability map.
{The overlapping area threshold in NMS is set to $0.68$, which means the new patch with previous selected patched having overlapping ratio larger than $0.68$ will be discarded.}
In this way, we can acquire $m=3m'$ discriminative patches for each whole slide image.
Lastly, we extract features $\mathbf{U}$ from the FConv layer of the localization network without activation operation as the discriminative instance representation.

\subsection{RMDL Network for Image-level Prediction}
In this subsection, we introduce our designed recalibrated multi-instance deep learning (RMDL) network for image-level prediction by considering the various influence of discriminative patches on the final image-level prediction process. 
As shown in Figure~\ref{fig:whole_architecture}, the RMDL network consists of three modules: local-global feature fusion, recalibration module, and multi-instance pooling.

\para{Local-global Feature Fusion.}\
Our goal is to predict the image label of WSI according to the features of the selected discriminative instances.
In order to eliminate the effects of the limited local information of each instance,
we aggregate the local and global features for the WSI classification task. 
In order to fulfill this requirement, we first use three consecutive fully-connected (FC) layers to extract different level features ($fc1$, $fc2$, and $fc3$) of input instances, as shown in local-global feature fusion module of Figure~\ref{fig:whole_architecture}.
Then we obtain two sub-global features of consecutive levels by applying max pooling operations on $fc1$ and $fc2$, respectively. 
The final global feature representation is generated by concatenating these two sub-global features.
We tile this global feature after each instance feature in $fc3$ to obtain the combined local and global features $\mathbf{H}$. 
Each $fc$ layer is followed by an instance normalization layer, a Leaky ReLU activation ($\alpha$ of 0.2), and a Dropout layer (rate of 0.5), except the last one with the $softmax$ activation function.
After this local and global feature fusion module, each instance representation is not limited to local feature but also has the ability to catch holistic image-level information.
This fusion of local and global feature can capture the inter-dependencies of different instances and make the generated importance coefficient more reliable.

\para{Instance Recalibration.}\
Each instance has a specific contribution to the final image-level prediction according to the inherent discriminability.
Therefore, we propose an instance recalibration module to recalibrate instance features according to the importance of each instance to the image label.
This mechanism could make the image-level prediction more robust even if there exist some irrelevant instances selected from the first stage.
In particular, we calculate the importance coefficient $\alpha$ for each instance from the combined feature $\mathbf{H}$ inspired by previous attention related works~\citep{lin2017structured,ilse2018attention}.
These coefficients identify the contribution of the corresponding instance to the final image-level prediction.
We calculate the importance coefficient $\alpha$ from the fused local and global feature $\mathbf{H}$ which contains image-level holistic information and could capture the inter-dependencies of different instances implicitly.
Suppose that the combined feature of the $i$-th instance can be represented as $\mathbf{h}_i$, the corresponding importance coefficient $\alpha_i$ can be formulated as follows.
\begin{equation}
	\alpha_i = \frac{exp(\mathbf{W}^T \mathbf{h}_i+b)}{\sum_{j=1}^{m}exp(\mathbf{W}^T \mathbf{h}_j+b)},\ \ \  i\in\{1,2,...,m\}
\end{equation}
where $m$ represents the number of instances in one WSI, $\mathbf{W} \in \mathbb{R}^{L \times 1 }$ and $b \in \mathbb{R}$ are the weights used for learning the importance, and $L$ is the dimension of feature $\mathbf{h}_i$.
Softmax operation is utilized to normalize the importance coefficient among all the instances.
The recalibrated instance feature $\hat{\mathbf{U}}$ is the element-wise multiplication of the coefficients and the original instance features $\mathbf{U}$.
Then the $i$-th recalibrated instance feature $\hat{\mathbf{u}}_i$ can be formulated as
\begin{equation}
	\hat{\mathbf{u}}_i = \alpha_i \cdot \mathbf{u}_i.
\end{equation}
Similar channel-based attention mechanism was also proposed in SENet \citep{hu2018squeeze}. However, our method is different from it in following aspects.
The SENet consists of SE blocks in order to strengthen the presentational power of its hierarchy features; while our method utilizes the attention mechanism in order to recalibrate the importance weights of instances to improve the image-level classification performance. Therefore, we apply such an importance recalibration mechanism on the instance level (2-D features [channel $\times$ instance]) rather than the channel level (3-D features [channel $\times$ width $\times$ height]). Another difference is that the local and global fusion features are utilized to calculate the recalibration weight in our method, while only the local features are utilized in SENet. 

As for implementation, we use a recalibration layer (\ie, FC layer) to squeeze the above concatenated local and global instance features into one dimension. 
The parameters $\mathbf{W}$ and $b$ can be represented by the weight and bias of the fully connected layer in this situation.
Then we use a softmax activation layer to normalize the generated statistics. 
We could employ more complicated layers to learn the coefficient of importance. 
However, we achieve satisfying results only with a simple recalibration layer in our experiment.
Our proposed recalibration module has good information aggregation ability resulting from aggregating the local and global features.
Furthermore, it can selectively emphasize informative instances and suppress other trivial ones.

To obtain the final image-level prediction, the multi-instance pooling maps the recalibrated instance feature $\hat{\mathbf{U}}$ to a global image-level feature ${\mathbf{z}}$. In our implementation, we use the average pooling layer to conduct a multi-instance pooling operation.
Finally, we further employ an $fc$ layer followed by a softmax activation to generate the image-level prediction.

\subsection{Training Procedure}
\label{sec:trainingprocedure}

We extract training patches to train the localization network from the level with a resolution of 0.5034 $\mu$m/pixel.
Since we extract the normal training patches from all slide images and the number of dysplasia WSIs is half less than cancer (as shown in Table~\ref{table:data_statistic}), data imbalance is evident when training the localization network.
To overcome this problem, we manually control the proportions of different training patches to be equal during the training of the localization network.
Specifically, we fetch the same number of normal, dysplasia, and cancer patches to form one batch during network optimization process.
To avoid overfitting, we also conduct the data augmentation strategy on-the-fly when training the localization network. 
The size of extracted training patches (\ie, $470\times 470$) is about $1.57$ times larger than the size of network input. We then randomly scale the extracted patches between $0.9$ and $1.1$, rotate it with an arbitrary angle, and flip it randomly.  
Finally, we crop the sub-patch with the size of $299 \times 299$ from the center of processed patch to train the network.
We use the trained network to infer the training samples and extract false positives to expand training samples for hard negative mining.
We extract $m'=100$ discriminative patches for normal, dysplasia and cancer types, respectively. So that we feed $m=300$ instances into the RMDL network.


\section{Experiments and Results}
\label{sec:experiments}

\subsection{Implementation details}
The proposed framework was implemented in python with Keras library~\citep{chollet2015keras} on a server equipped with four NVIDIA TITAN Xp GPUs. 
{
	In order to enhance the training efficiency, we used the pre-trained Inception ResNet\_v2 model (trained on the ImageNet dataset \citep{russakovsky2015imagenet}) to initialize the weights of the localization network. 
}
The localization network was trained with Adam optimizer~\citep{kingma2014adam} ( $\beta_1$, $\beta_2$, and decay were set as 0.9, 0.999, and 0.001 respectively).
We trained the localization network with four GPUs and the batch size was $80$.
The learning rate was set as $1e-4$ for network training and multiplied $0.9$ every $2000$ iterations.
Since reading WSI into the memory is quite slow, we utilized the Asynchronous Sample Prefetching mode in \cite{lin2018scannet} to speed up the whole workflow.
For training the RMDL network, we also used Adam optimizer with the same parameter setting with the localization network training except for the initial learning rate of $1e-3$. 
The learning rate was reduced by multiplying a factor of 0.8 every 140 iterations for totally 7000 iterations. 
We trained the RMDL network only with one GPU and a batch size of 256.
Instance feature permutation was utilized to increase training samples, avoid over-fitting, and guarantee the permutation-invariant attribute for the MIL model.

\subsection{Evaluation Metrics}
In our experiments, accuracy and \textit{average classification score} were used for WSI classification evaluation. 
Accuracy is a common quantitative evaluation metric for classification. 
WSIs are classified into three classes: normal, dysplasia, and cancer in our scenario. 
These classes have hierarchical differences, and the risk of misclassification is different.  
For a normal image, it is more severe for the network to classify it as cancer than dysplasia.
While for a cancer image, it is more severe if the network classifies it as normal but less for dysplasia. 
Therefore, we further designed the \textit{average classification score} to evaluate the classification performance following the Nuclear Atypia evaluation in MITOS-ATYPIA-14 Challenge\footnote{https://mitos-atypia-14.grand-challenge.org/evaluation-metrics/}.
Here, if we classify one image correctly, we will get the standard 2 points. If a prediction is wrong and the absolute value of the difference between annotated and predicted gastric cancer grade is $1$,  then we will get $1$ point. Otherwise, we will get $-1$ point.
We normalized this score to get \textit{average classification score} by dividing the total point when all the testing slides were predicted correctly.
Note that even if some methods have the same accuracy, they may get different \textit{average classification score}. This score, as a soft quantitative measure, is a more proper evaluation metric in our problem.

\subsection{Qualitative Evaluation of Discriminative Instance Selection}
\begin{figure*}[t]
	\centering
	\includegraphics[width=0.75\linewidth]{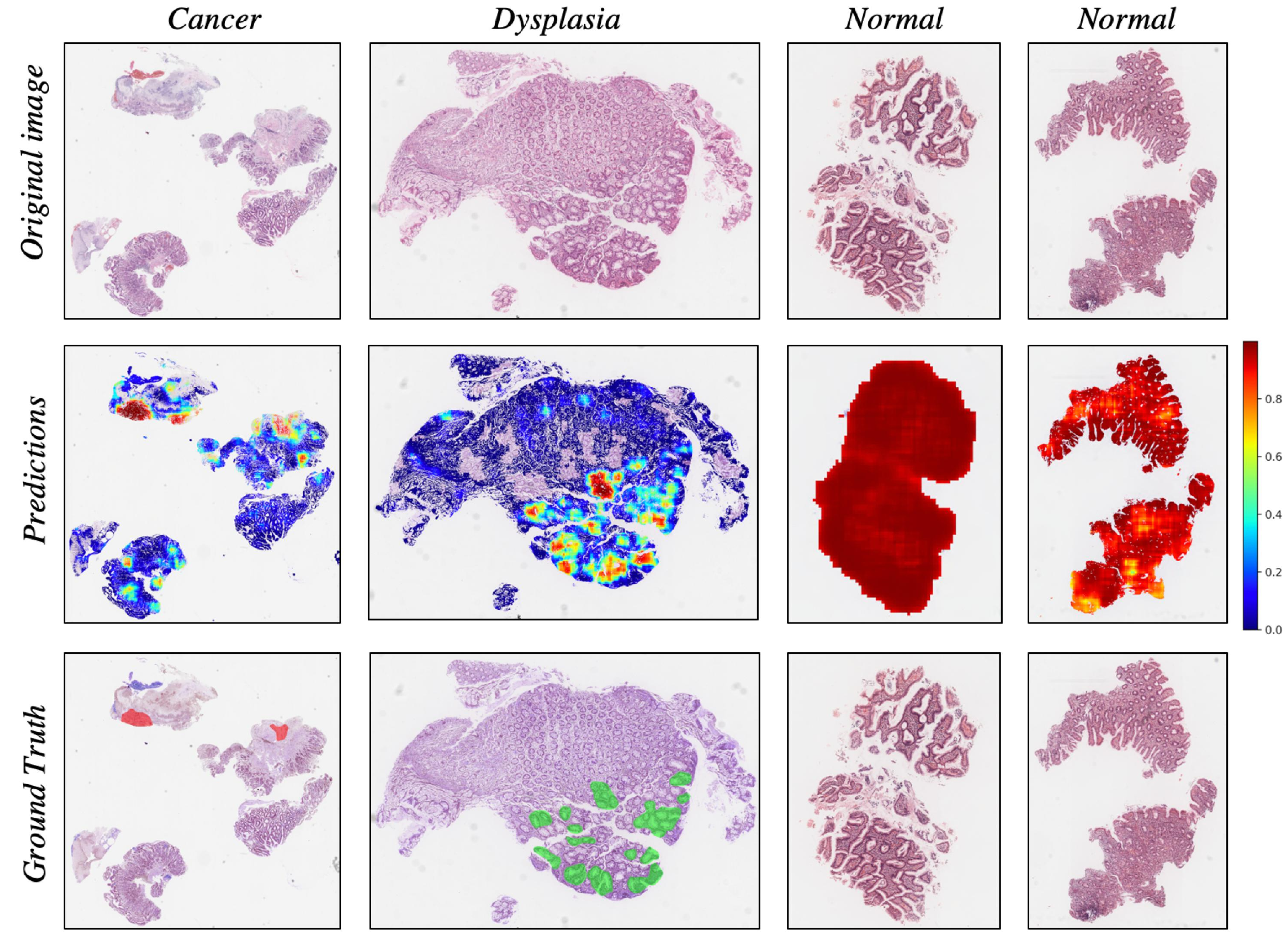}
	\caption{Abnormal region detection results (Best viewed in color).
		The color bar reflects the probability values for the corresponding type in the prediction row.
		The \textit{red} and \textit{green} annotations in the \textit{Ground Truth} row represents the cancer and dysplasia regions, respectively, while the remaining regions are normal.
	}
	\label{fig:localization_evaluation}
\end{figure*}

To demonstrate the effectiveness of our proposed discriminative instance selection procedure, we visualized some qualitative results of abnormal detection
in Figure~\ref{fig:localization_evaluation}.
{Note that the first and second columns are cancer and Dysplasia slides, while the last two columns show the normal cases.}
Also, the probability values in the \textit{Prediction} row are corresponding to the WSI types.
We can see that our localization network can accurately locate cancer and dysplasia regions from the whole slide image, although these regions are dispersed and some of them are tiny.
Due to the complex texture and structural diversity of the whole slide images, our localization network generated some false positive small abnormal regions.
However, our discriminative instance selection procedure can remove these outliers due to the lower probability of these regions.
These visualization results show that our first stage would effectively extract the most discriminative regions for the future image-level prediction.

\subsection{Quantitative Comparison with Other Methods}
We evaluated the effectiveness of the proposed RMDL network by comparing it with other state-of-the-art methods.
The comparison results are shown in Table~\ref{table:results_others}.

\begin{table}[!ht]
	\begin{center}
		\caption{Comparison with other methods.}
		\label{table:results_others}
		\resizebox{0.4\textwidth}{!}{
		\begin{tabular}{l|cc}
			\hline    
			\textbf{Method}  & \textbf{Average score} & \textbf{Accuracy (\%)} \T \B \\
			\hline    
			CNN-Vote-LR \citep{Hou}      & 0.533  & 45.5 \T\B \\
			CNN-Vote-SVM \citep{Hou}     & 0.695 &  46.0 \B \\
			MIMLNN \citep{mercan2018multi} & 0.803 &74.5 \B \\
			MI-NET-RC \citep{wang2018revisiting}  & 0.825 & 76.0 \B\\
			CNN-DesignFeat-RF \citep{wang2016deep} & 0.803 & 77.5 \B \\ 
			MI-NET-DS \citep{wang2018revisiting}  & 0.875 & 78.0 \B\\
			MAXMIN-Layer \citep{courtiol2018classification} & 0.893 & 79.5 \B\\
			Attention-MIP \citep{ilse2018attention} & 0.875    & 82.0 \B \\ 
			MISVM \citep{andrews2003support} & 0.908 & 82.5 \B \\ 
			\textbf{RMDL (Ours)} & \textbf{0.923} & \textbf{86.5}  \B \\
			\hline
		\end{tabular}}
	\end{center}
\end{table}

The comparison results show the performance of each method based on handcrafted features or CNN-based features.
We first show the results on the handcrafted features used in~\cite{Hou}.
To obtain the handcrafted features, we used the localization network to generate the labels of all patches extracted from the slide image and then counted the numbers of different predicted labels to form a prediction histogram.  
We then used a multi-class logistic regression (CNN-Vote-LR) and SVM (CNN-Vote-SVM) algorithm to predict the image-level labels utilizing the normalized count numbers.
It is observed that the performance of these two decision fusion methods is not very competitive. 
One of the possible reasons for the poor results is that the discriminative information is suppressed by the normal patch proportion, due to the small abnormal regions in our gastric dataset.
Another way to extract the hand-crafted features is to extract geometrical and morphological features from predicted probability maps and use random forest classifier to classify the WSIs (referred to CNN-DesignFeat-RF).  
It is observed that our method outperforms CNN-DesignFeat-RF on the whole slide image classification task about $12\%$ accuracy.
Then, based on the CNN extracted feature, we conducted experiments to show the effectiveness of other methods.
We compared our RMDL network with other multi-instance learning methods with the same extracted discriminative instances.
Among them, the MAXMIN-Layer~\citep{courtiol2018classification} combined top positive and negative instances for image-level prediction. 
In~\cite{wang2018revisiting}, two multi-instance deep learning method: MI-NET-DS and MI-NET-RC were proposed with deep supervision and residual connections, respectively.
We also compared our RMDL network with Attention-MIP~\citep{ilse2018attention}, which was proposed to solve the multi-instance classification problem in an attention mechanism.
Besides these neural network based multi-instance learning methods, we further compared with MISVM and MIMLNN. 
The MISVM~\citep{andrews2003support} was a modified SVM to handle multiple instance learning problems. 
We utilized a one vs. rest strategy for each class to implement multi-class classification in this method.
MIMLNN~\citep{mercan2018multi} used a sum-of-squares error function to estimate the weights of a linear classifier to solve a multi-instance multi-label problem.
We can regard our problem as a multi-label (cancer and dysplasia labels) problem so that it can be solved by this method.
From the comparison in Table~\ref{table:results_others}, we can see that our RMDL network achieves the best performance than other multi-instance deep learning methods, showing the effectiveness of our proposed instance recalibration module. 
The attention-based method  MAXMIN-Layer also produces competitive results. 
Specifically, our approach is better than Attention-MIP, as we integrate the local and global feature to capture the inter-dependencies of the different instance to learn the importance coefficient better.

\begin{figure*}[!thb]
	\centering 
	\begin{subfigure}{0.25\linewidth}
		\centering
		\includegraphics[width=\linewidth]{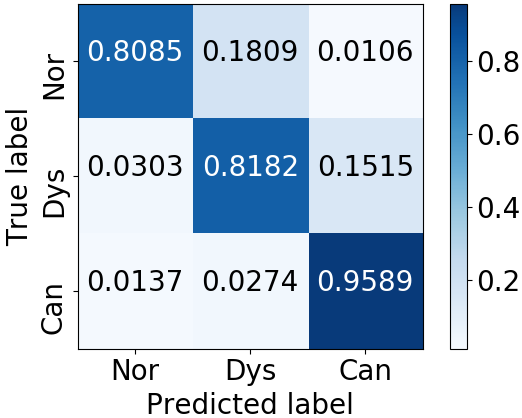}
		\caption{{RMDL.}}
		\label{fig:1}
	\end{subfigure}
	\begin{subfigure}{0.25\linewidth}
		\centering
		\includegraphics[width=\linewidth]{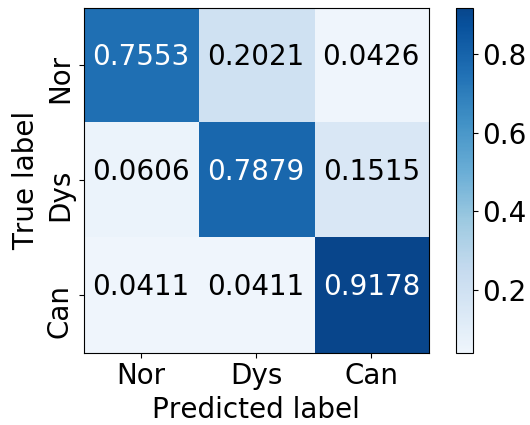}
		\caption{{Attention-MIP.}}
		\label{fig:2}
	\end{subfigure}
	\begin{subfigure}{0.25\linewidth}
		\centering
		\includegraphics[width=\linewidth]{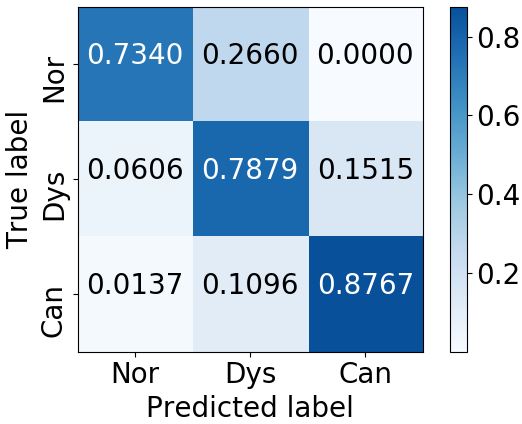}
		\caption{{MAXMIN-Layer.}}
		\label{fig:3}
	\end{subfigure}
	
	\caption{
		{Confusion matrix analysis. ‘Nor’, 'Dys', and 'Can'  are the abbreviations of 'Normal', 'Dysplasia', and 'Cancer'. }
	}
	\label{fig:cm}
\end{figure*}

\begin{figure*}[!thb]
	\centering 
	\begin{subfigure}{0.25\linewidth}
		\centering
		\includegraphics[width=\linewidth]{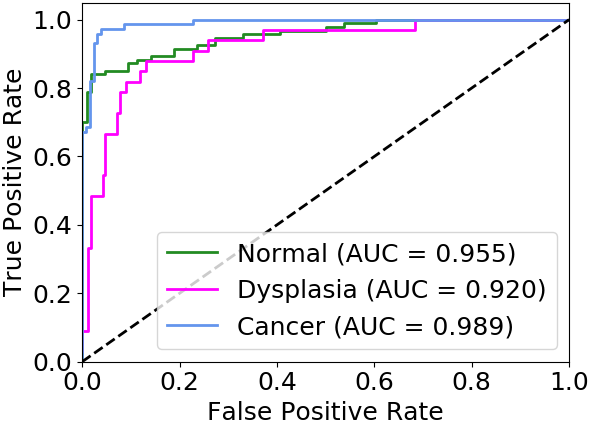}
		\caption{{RMDL.}}
		\label{fig:roc1}
	\end{subfigure}
	\begin{subfigure}{0.25\linewidth}
		\centering
		\includegraphics[width=\linewidth]{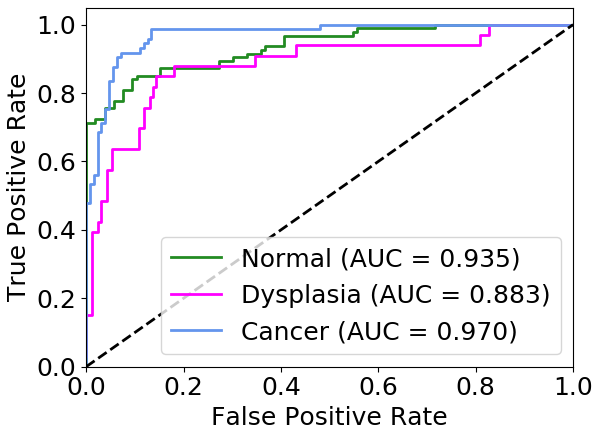}
		\caption{{Attention-MIP.}}
		\label{fig:roc2}
	\end{subfigure}
	\begin{subfigure}{0.25\linewidth}
		\centering
		\includegraphics[width=\linewidth]{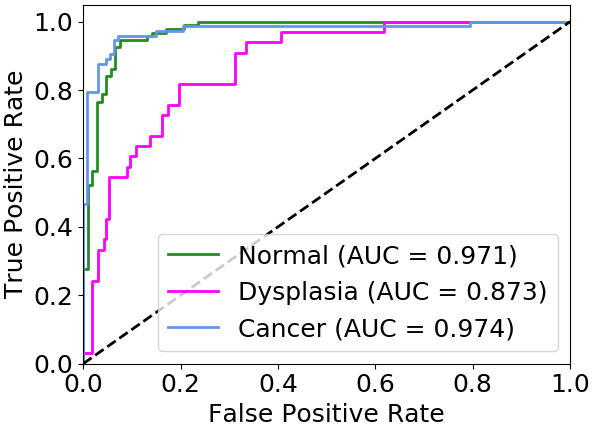}
		\caption{{MAXMIN-Layer.}}
		\label{fig:roc3}
	\end{subfigure}
	\caption{
		{The ROC curve analysis. The True positive Rate and False Positive Rate are calculated through a one-vs.-rest strategy based on the classification results. }
	}
	\label{fig:roc}
\end{figure*}

In order to compare the classification performance for different classes, we show the confusion matrix and ROC curve of the proposed method RMDL, Attention-MIP, and MAXMIN-Layer methods in Figure~\ref{fig:cm} and Figure~\ref{fig:roc}.
The ROC curves were drawn through a one-vs.-rest strategy.
As for the confusion matrix, our proposed method achieves the highest accuracy for each image type.
Moreover, the AUC values of dysplasia and cancer images of the proposed RMDL outperform the other two methods.

\subsection{Analysis of Our Method}
{In this subsection, we extensively analyze the behavior of different components in our designed method.}
To discover the vital elements in the success of our proposed RMDL network, we conducted ablation studies by removing local and global feature fusion and instance recalibration module.
The experimental results are shown in Table~\ref{table:results_abalation}.
The baseline model only contains a multi-instance pooling (MIP) module and a classification layer (fully connected layer).
The IR module and LG module represents the instance recalibration module and local-global feature fusion module, respectively.
As we can see from the Table~\ref{table:results_abalation}, both two modules contribute a lot for high classification accuracy. 
The IR and LG modules have 2.5\% and 4.5\% promotions on the classification accuracy, respectively.
And the average scores also improve a lot with the equipment of multi-instance pooling and local-global feature fusion modules. By combining both modules, we further boost the final image-level classification accuracy and average score.
\begin{table}[!h]
	\begin{center}
		\caption{Ablation studies of RMDL network on multi-instance pooling (MIP), instance recalibration module (IR), and local-global feature fusion module (LG).}
		\label{table:results_abalation}
		\resizebox{0.4\textwidth}{!}{
		\begin{tabular}{ccc|c|c}
			\hline    
			\textbf{MIP} & \textbf{IR} & \textbf{LG} & \textbf{Average score} & \textbf{Accuracy (\%)} \T\B\\
			\hline    
			\checkmark    &         &            & 0.858 & 79.5\T\B \\
			\checkmark    &   \checkmark    &       & 0.875  & 82.0\B \\
			\checkmark && \checkmark & 0.910 & 84.0\B \\
			\checkmark    &    \checkmark    &        \checkmark            & \textbf{0.923} & \textbf{86.5}\B \\ 
			\hline
		\end{tabular}}
	\end{center}
\end{table}


We also conducted experiments to analyze the influence of different numbers $m'$ of selected discriminative instances, as shown in Table~\ref{table:feature-number}.
We can observe that there is a slight fluctuation of the classification accuracy with different discriminative instance number setting.
This comparison further demonstrates that our RMDL network is very robust and not sensitive to the discriminative instance numbers as long as the most important instances are selected.

\begin{table}[!h]
	\begin{center}
		\caption{Analysis of different number of discriminative instances.}
		\label{table:feature-number}
		\resizebox{0.3\textwidth}{!}{
		\begin{tabular}{c|c|c}
			\hline    
			$m'$  & \textbf{Average score} & \textbf{Accuracy (\%)} \T \B\\
			\hline    
			30                             & 0.910 & 85.0 \T\B \\ 
			50                            & 0.923  & 85.5 \B \\
			100                            & 0.923 & 86.5  \B\\
			\hline
		\end{tabular}}
	\end{center}
\end{table}



{Finally, the computation cost of the method designed for WSI is also attractive.
	We report the average computation cost of Testing part on the collected WSGI dataset and the computation cost of the smallest and largest slides in Table \ref{table:cost}.}
We split the computation cost into two parts: instance selection and image-level prediction. On average, the total computation cost is $93.79$ seconds which is around $1.5$ minutes for the images with {an average size of $61858\times43834$ $pixel^2$.} And the time-consuming is positive related to the image size. We observe that the instance selection step occupies the maximum perception of total computation cost and the final image-level prediction only cost about $0.01$ seconds.
In the future, we would study knowledge distillation to find a light-weight model for faster inference of the instance selection step.
\begin{table}[!h]
	\begin{center}
		\caption{Computation cost analysis for the Testing part. We show the average computation cost and the cost of the smallest and largest slides on two steps (instance selection (IS) and prediction (P)).}
		\label{table:cost}
		\resizebox{0.45\textwidth}{!}{
		\begin{tabular}{c|c|c|c}
			\hline    
			\textbf{Image size ($pixel^2$)}  & \textbf{IS ($s$)}  & \textbf{P ($s$)} & \textbf{Total ($s$)} \T \B\\
			\hline    
			Average ($61858\times43834$) & 93.78 & 0.01 & 93.79 \T \B \\
			Small slide ($9019\times31880$) & 23.60 & 0.01 & 23.61 \T \B \\
			Large slide ($171312\times82473$) & 393.87 & 0.01 & 393.88 \T  \\
			\hline
		\end{tabular}}
	\end{center}
\end{table}

\section{Discussion}
\label{sec:discussion}

In this work, we presented a novel recalibrated multi-instance deep learning based framework for the whole slide gastric image classification.
{This is an important problem in medical image analysis, since histopathology image analysis is the gold standard for cancer diagnosis.}
The accurate and automated computer-aided whole slide classification could mitigate the lack of medical resources and time cost of manual diagnosis in clinical practice.

However, it remains a challenging problem of automatic diagnosis for gastric cancer based on the whole slide histopathology images due to the large image scale.
Some weakly supervised methods were proposed to tackle this classification problem with only image-level labels~\citep{xu2014weakly, Hou, wang2018weakly, courtiol2018classification, mercan2018multi}.
These methods usually made strong assumptions (\eg, most of the patches are with the same label with the WSI) and cannot generalize well to specific tasks (\eg, disproportion dataset in our work). 
Due to the large image size, the image-level labels with a small proportion of lesion makes it difficult for automated diagnosis, especially for the disproportion dataset.
In the context of the medical image analysis, more detailed pixel-level annotations are in demand for reliable results.
For the disproportion dataset used in this paper, we optimized a high precision localization network and recalibrated multi-instance learning to solve the WSI classification problem.
However, the pixel-level annotations are subjective, time-consuming, and expensive to obtain.
{Generally, it takes around half an hour for experienced pathologists to annotate one whole slide image exhaustively.}
To reduce the annotation effort while maintaining the method accuracy, one possible solution is utilizing the semi-supervised methods~\citep{weston2012deep,tarvainen2017mean}.
These methods can leverage both a limited amount of labeled and an arbitrary amount of unlabeled data.
For example, \cite{tarvainen2017mean} designed a teacher-student training strategy and achieved state-of-the-art results using semi-supervised learning for computer vision. The principle of this method was to encourage the consistent prediction of the network for the same input data under different regularizations.
{Some recent works~\citep{gu2017semi,sedai2017semi,bai2017semi,yu2019uncertainty} also introduced semi-supervised methods into medical image analysis community.}
In the future work, it will be very promising to extend our method to train the localization network with semi-supervised learning techniques by utilizing more unlabeled datasets.

The experimental results demonstrated the effectiveness of our proposed recalibrated multi-instance deep learning method. 
The essence of our RMDL is the instance recalibration module, which can automatically figure out the crucial instances for image-level prediction.
{This recalibration process is similar to the attention-based method that makes us pay selective attention to different aspects.}
This kind of attention model has been used widely in nature image processing, such as image classification~\citep{NIPS2014_5542, xiao2015application, wang2017residual,jetley2018learn} and semantic image segmentation~\citep{chen2016attention,ren2017end}.
Similar attention methods were also used for solving medical image analysis problems recently.
For example, \cite{oktay2018attention} introduced attention gate model for automatic pancreas segmentation from CT images. 
\cite{bychkov2018deep} proposed a convolutional and recurrent architecture to predict five-year disease specific survival.
Our work and these studies demonstrated the potential application of attention mechanism in medical image analysis field.
{The computation cost of proposed method is still required to be improved. There exists a trade-off between precision and speed for the localization network architecture. 
	Therefore, exploring more light-weight and speedy network architecture through knowledge distillation \citep{hinton2015distilling,wang2019Private} will be a feasible solution to speed up the detection process in the future.
}


Furthermore, due to the time and labor effort of annotating whole slide images, there is no public benchmark dataset for gastric histopathology image classification.
We built a sizeable whole slide gastric histopathology image dataset (WSGI) with pixel-level annotations.  
Our proposed method achieved the best performance on the WSGI dataset compared with other methods. 
In the future, we will test our methods on other public datasets to evaluate the generalization capability of our method.
It should be noted that one of the limitations of this work is that we trained the two-stage framework separately due to the technical issues (\eg, limited GPU memory). 
In this way, the extracted instance features by the localization network may be not the best choice for RMDL network. In the future, we will explore a parallel model technique to train our RMDL network in an end-to-end fashion. 

\section{Conclusions}\label{sec:conclusions}

In this work, we proposed a two-stage pipeline for whole slide gastric image classification based on discriminative patch selection and image-level label prediction.
The main contribution of our work is the proposed recalibrated multi-instance deep learning network for adaptively aggregating the patch information to image-level prediction.
The proposed RMDL network considered the different impacts of each instance to the final image label decision and thus recalibrated each instance feature according to the learned importance coefficient.
In the evaluation process, our method achieves the 86.5\% classification accuracy on the testing images, and the proposed RMDL network outperforms the state-of-the-art multi-instance learning method by a large margin (4.0\%). 
{Further investigations, including exploring the semi-supervised method to reduce the annotation effort and designing model compression methods to speed up the detection process, are necessary.}

\section*{Acknowledgment}
This work was supported in part by the Hong Kong Innovation and Technology Commission through the ITF ITSP Tier 2 Platform Scheme under Project ITS/426/17FP, and in part by Shenzhen Science and Technology Program under No.JCYJ20180507182410327.

\bibliography{ref}

\end{document}